\documentclass[conference]{IEEEtran}

\IEEEoverridecommandlockouts
\usepackage{cite}
\usepackage{amsmath,amssymb,amsfonts}
\usepackage{algorithmic}
\usepackage{graphicx}
\usepackage{textcomp}
\usepackage{xcolor}
\usepackage{soul}
\usepackage{breqn}
\usepackage{flushend}
 
\newcommand{\Lagr}{\mathop{\mathcal{L}}}
\usepackage{tabularx}
\usepackage[subrefformat=parens]{subcaption}
\usepackage{comment}

\def\BibTeX{{\rm B\kern-.05em{\sc i\kern-.025em b}\kern-.08em
    T\kern-.1667em\lower.7ex\hbox{E}\kern-.125emX}}

\title{Privacy-Preserving Taxi-Demand Prediction Using Federated Learning}

\makeatletter
\newcommand{\linebreakand}{%
  \end{@IEEEauthorhalign}
  \hfill\mbox{}\par
  \mbox{}\hfill\begin{@IEEEauthorhalign}
}
\makeatother

\author{\IEEEauthorblockN{Yumeki Goto}
\IEEEauthorblockA{Osaka University, Japan \\
y-goto@ist.osaka-u.ac.jp}
\and
\IEEEauthorblockN{Tomoya Matsumoto}
\IEEEauthorblockA{Osaka University, Japan \\
t-matsumoto@ist.osaka-u.ac.jp}
\and
\IEEEauthorblockN{Hamada Rizk}
\IEEEauthorblockA{Osaka University, Japan \\
Tanta University, Egypt \\
hamada\_rizk@ist.osaka-u.ac.jp} 
\linebreakand
\IEEEauthorblockN{Naoto Yanai}
\IEEEauthorblockA{Osaka University, Japan \\
yanai@ist.osaka-u.ac.jp}
\and
\IEEEauthorblockN{Hirozumi Yamaguchi}
\IEEEauthorblockA{Osaka University, Japan \\
h-yamagu@ist.osaka-u.ac.jp}
}

\begin{document}

\maketitle

\begin{abstract}
Taxi-demand prediction is an important application of machine learning that enables taxi-providing facilities to optimize their operations and city planners to improve transportation infrastructure and services. However, the use of sensitive data in these systems raises concerns about privacy and security. 
In this paper, we propose the use of federated learning for taxi-demand prediction that allows multiple parties to train a machine learning model on their own data while keeping the data private and secure. 
This can enable organizations to build models on data they otherwise would not be able to access.
Evaluation with real-world data collected from 16 taxi service providers in Japan over a period of six months showed that the proposed system can predict the demand level accurately within 1\% error compared to a single model trained with integrated data. 
\end{abstract}


\section{Introduction}

The utilization of spatio-temporal location data has immense potential to enhance the availability and improvement of various services, especially data-driven approaches, which can train intelligent models in different domains, such as transportation, urban planning, and emergency management. One such service, taxi transportation, is a critical component of modern urban transportation systems, providing convenient and efficient transportation to a wide range of passengers. However, there is often a mismatch between the supply of taxis and passenger demand, leading to decreased profits for taxi providers due to increased cruising times, fuel consumption, and longer wait times for customers.

To address this issue, taxi-demand prediction systems have been proposed that utilize data-driven approaches to predict taxi demand and optimize dispatch processes~\cite{Dynamic_Taxi_Ride-Sharing, Understanding_Taxi_Service_Strategies}. Machine or deep learning models are trained with real customer mobility data to forecast future taxi demand in a specific geographic area. This training data includes pickup and drop-off locations, routes taken, and timing information of customers. 
However, sharing such trajectory data raises significant privacy concerns as it could reveal intimate personal details, such as individuals' whereabouts, movement patterns, and even their religious, political, or sexual convictions, through the prediction of Points of Interest (POI) using mapping data and coordinates. facilities may have different legal and regulatory requirements that they need to comply with. These requirements can vary between countries and regions and need to be considered when working with data from different facilities.

Various privacy-preserving methods \cite{4497446,4417165,10.1145/1869790.1869846,10.1145/2484838.2484846,10.1145/3423165,ohno2023privacy,rttMDM} have been proposed to address privacy concerns associated with personal data. These methods aim to protect the privacy of individuals by anonymizing the data before sharing it. Differential privacy is a method that introduces randomness into data, making it difficult for an attacker to determine the identity of individuals \cite{10.1145/2484838.2484846}. K-anonymity groups individuals into groups with similar characteristics, making it difficult to determine the identity of any individual\cite{4497446}. L-diversity and t-closeness are other privacy-preserving methods that generalize data to prevent sensitive information disclosure\cite{8329504,10.1145/1217299.1217302, li2006t}. Secure computation allows for the computation of a function on private data without revealing it \cite{arxiv.1610.05492,8859260,QI2021328}. While these methods can protect privacy, they can also result in a loss of data quality and quantity, negatively impacting the performance of the service (e.g., the prediction accuracy of taxi demand). Thus, it is important to weigh the trade-off between privacy and performance when choosing a privacy-preserving method.

In this paper, we propose a novel taxi-demand prediction system that prioritizes customer privacy and builds the model without necessitating sharing data. This can be achieved by employing federated learning that allows multiple parties to train a machine learning model on their own data while keeping the data private and secure. In the context of taxi-demand prediction, this could be useful because it allows multiple facilities (e.g., taxi service providers) to collaborate on building a demand prediction model without sharing their proprietary data with each other. This can lead to more accurate predictions, as the model is able to learn from a larger and more diverse dataset.

However, the application of federated learning in this context faces a  generalization problem as the local models are trained with absolute latitude-longitude values associated with each facility's data. 
The use of absolute latitude-longitude values may exhibit \textit{region-dependence characteristics} that affect the generalization ability and convergence of the global prediction model.
To address this challenge, the system incorporates a number of techniques to encode the absolute latitude-longitude values into a region-independent space, making the model more versatile and applicable to different geographical areas.

{\color{blue}

\begin{figure}[!t]
\centering
   \begin{minipage}[b]{0.87\linewidth}
\includegraphics[width=\linewidth]{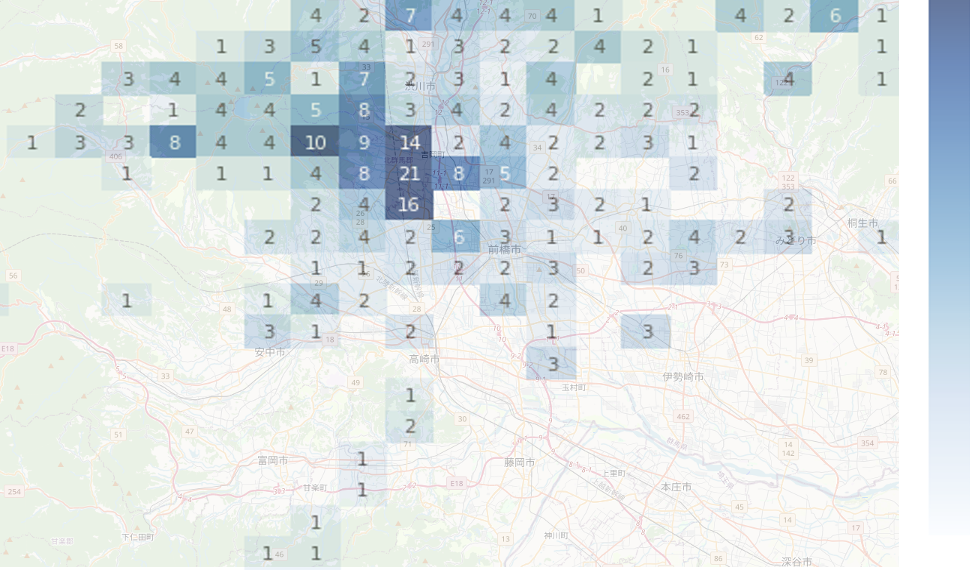}
\caption{An example of how the taxi demand is biased toward no or low level in an area of one facility. X-Y are the lat-long values and the boxes represent the number of taxi requests in this spot at a specific time.}
\label{fig:heatmap}
  \end{minipage}
\end{figure}


}

The proposed system was subjected to a rigorous evaluation using real-world data gathered from 16 taxi service providers in Japan. The data was collected over a six-month period and employed to evaluate the system's effectiveness in maintaining prediction performance while preserving passenger privacy. The results obtained from the evaluation confirm that the proposed system, which utilizes federated learning and associated modules, achieves a comparable accuracy level with a negligible reduction of less than 1\% in accuracy compared to non-federated learning approaches that require sharing of customer data among facilities.

The rest of the paper is organized as follows: 
Section~\ref{sec:related_works} contains related works. 
Section~\ref{sec:TheSystemDetails} explains our federated learning system for taxi-demand prediction in detail. 
Section~\ref{sec:evaluation} discusses evaluations of the system. Finally, the conclusions are discussed in Section~\ref{sec:conclution}.

\section{Related Work} \label{sec:related_works}

This section describes taxi-demand prediction and privacy-preserving machine learning, including federated learning in spatiotemporal data and several privacy-preserving notions, as related works.

\subsection{Taxi-Demand Prediction} 
The prediction of taxi demand has recently garnered considerable attention, owing to the abundance of large-scale spatiotemporal data that facilitates the training of deep neural networks, such as Convolutional Neural Networks (CNNs) and Long Short-Term Memory (LSTM) networks.

Recent studies have leveraged both spatial and temporal characteristics to predict taxi demand with greater accuracy. For example, \cite{Yao_Wu_Ke_Tang_Jia_Lu_Gong_Ye_Li_2018} employs a CNN to capture spatial features and an LSTM to capture temporal features, resulting in improved accuracy compared to methods that only consider semantic, spatial, or temporal information. \cite{9172100} recognizes the existence of spatiotemporal correlations between pick-up and drop-off locations and proposes a taxi-demand prediction model using multitask learning, which predicts both pick-up and drop-off locations as interrelated tasks. This approach leads to more accurate prediction results.


Other studies have focused on accounting for the heterogeneity of taxi demand across regions. \cite{9439926} clusters taxi-demand data and trains region-specific models to predict demand, taking into account the unique distribution and temporal variability of demand in each region. While these machine learning-based methods have shown promising results when applied to spatiotemporal data, they do not consider privacy threats associated with sharing users's data, even anonymized. The methods proposed in \cite{ourMDMpaper,sharingHamada} represent  groundbreaking approaches to sharing synthetic versions of data by utilizing generative adversarial networks, thereby enabling secure data publication.

\textit{In contrast, our proposed system evaluates the accuracy of taxi-demand prediction while preserving privacy. The system uses federated learning to avoid sharing sensitive customer data. }

\subsection{Privacy-Preserving Machine Learning} 


The main motivation for federated learning in spatiotemporal data is for privacy-preserving on heterogeneous data~\cite{wen2023asurveyonfederated} that may cause a model drift problem for conventional training algorithms. 
Federated learning in spatiotemporal data is often discussed in actual application environments, i.e., urban~\cite{li2022federatedmeta-learning,yuan2022fedstn}, renewable energy~\cite{li2022privacy-preserving}, and robotics~\cite{majcherczyk2021flow-fl}. 
In this paper, we discuss taxi-demand prediction as an application environment different from the above existing works. 


The most popular approach for privacy-preserving machine learning is differential privacy~\cite{Dwork06} which provides theoretical security. 
Differential privacy is used for gradient computation~\cite{abadi2016deeplearning}, and it can theoretically prevent data recovery~\cite{yeom2018privacyrisk}. 
There are results on differential privacy of federated learning~\cite{wei2020federated} and developments of libraries~\cite{yousefpour2021opacus,papernot2019tensorflow,prediger2022d3p}. 
However, differential privacy deteriorates accuracy significantly. 

Another approach for providing privacy is to achieve definitions such as k-anonymity~\cite{sweeney2002k,lefevre2006mondrian} and l-diversity~\cite{machanavajjhala2007diversity}. 
The k-anonymity requires each record to share the same values with at least k-1 other records in the dataset while the l-diversity requires each equivalence class to contain at least l sensitivities. 
Similar to differential privacy, accuracies of machine learning models based on these notions deteriorate~\cite{khan2022privacy,SLIJEPCEVIC2021k-anonymityinPractice}.

\section{The System Details} \label{sec:TheSystemDetails}

This section describes the proposed system in detail. 
The virtual gridding module and its resultant taxi-demand prediction model are first described. 
Then, the federated learning 
is described.

\subsection{The Virtual Gridding Module} 

The Virtual Gridding module is a crucial component that operates during both the online and offline phases of the system. In the offline phase, the module processes historical trajectory data to construct a comprehensive demand profile for the city. This profile is then used to train the machine learning models that power the demand prediction functionality of the system. The module achieves this by transforming the raw trajectory data collected from taxi drivers into a more manageable and interpretable format.

To accomplish this, the module creates a virtual grid, dividing the city map into evenly spaced grid cells that correspond to specific locations. By tracking the number of pick-up and drop-off events within each cell during a specified time-slot, the module accurately calculates the total demand events for each area. This approach enables the system to provide a high-level overview of the taxi demand across various regions of the city. The resulting demand patterns can then be leveraged to train machine learning models for predicting the number of demand events accurately in different cells. Furthermore, the grid-based visualization of the demand patterns can be used to identify areas of high or low demand quickly.

During the online phase, this module converts any latitude and longitude coordinate into the corresponding grid cell in real-time. This cell ID can be fed into the trained demand prediction model to make accurate real-time predictions
ensuring that the system has access to the most recent demand information.

\begin{figure}[!t]
\centering
   \begin{minipage}[b]{0.87\linewidth}
    \includegraphics[width=\linewidth,]{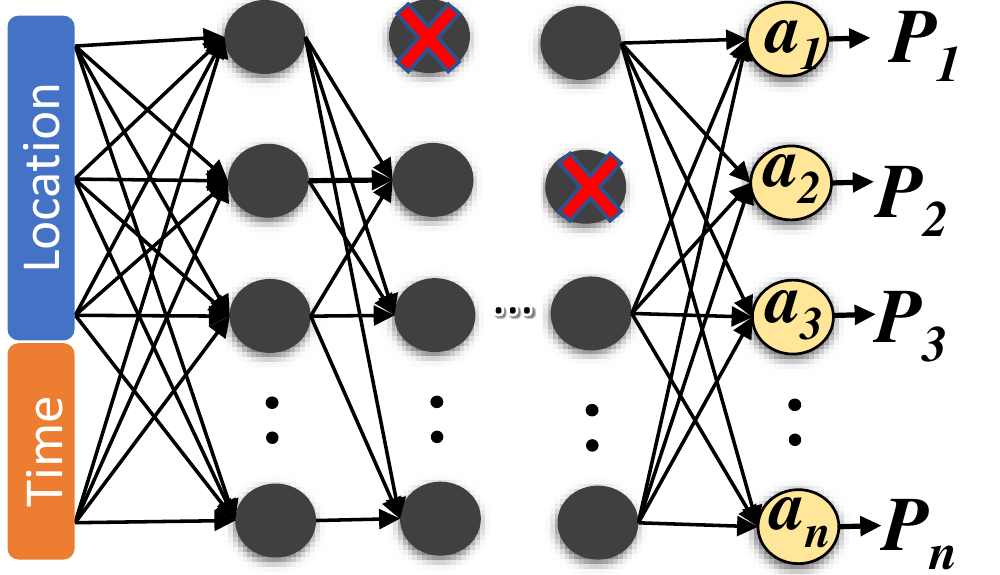}
\caption{Neural network structure for the taxi-demand prediction model. 
}
\label{fig:network}
  \end{minipage}
  \vspace{-0.5cm}
\end{figure}

\subsection{Taxi-Demand Prediction Model} 
This module is responsible for leveraging the input features ($c$) to train a deep localization model and find its optimal parameters. The trained model is used during the online phase by the \textit{Demand Predictor} module to provide an estimate of the taxi-demand. A deep fully-connected neural network is adopted here due to its representational ability, which allows the learning of complex patterns. 

\subsubsection{The Network Architecture} 
Fig.~\ref{fig:network} shows our deep network structure. We construct a deep fully connected neural network consisting of cascaded hidden layers of nonlinear processing neurons. 
Specifically, we use the hyperbolic tangent function (tanh) as the activation function for the hidden layers due to its non-linearity,  differentiability (i.e., having stronger gradients and avoiding bias in the gradients), and consideration of negative and positive inputs~\cite{lecun2012efficient}. 
The input layer of the network is the cell id and the timestamp.
 The output layer consists of a number of neurons corresponding to the number of taxi-demand levels in the data. This network is trained to operate as a  multinomial  (multi-class) classifier by leveraging a softmax activation function in the output layer. This leads to a probability distribution over the demand levels given a spatiotemporal input (cell location and time).

More formally, the input feature vector $c_i$ = $(c_{i1}, c_{i2}, ..c_{ik})$ of length $k$, the corresponding discrete outputs (i.e logits) $c_i$ is $a_i = (a_{i1},a_{i2}, ..,a_{in})$ capture the score for each demand level from the possible $n$ total taxi-demand levels to be the estimated level. The softmax function converts the logit score $a_{ij}$ (for sample $i$ to be at demand level $j$) into a probability as:
            \begin{equation}
            p(a_{ij})= \frac{e^{a_{ij}}}{\sum_{j=1}^{j=q}{e^{a_{ij}}}}
            \end{equation}
This module is  responsible for leveraging the input features ($c$) to train a deep localization model and find its optimal parameters. The trained model is used during the online phase by the \textit{Demand Predictor} module to provide an estimate of the taxi-demand. A deep fully-connected neural network is adopted here due to its representational ability, which allows the learning of complex patterns. 


            

\subsubsection{Training} 
During the training  phase, the ground-truth probability label vector of demand $P(a_i) = [p(a_{i1}), p(a_{i2})...p(a_{in})]$ is formalized using one-hot-encoding. 
This encoding has a probability of one for the correct demand levels and zeros for others. 

The model is trained using the Adaptive Moment Estimation (Adam optimizer~\cite{kingma2014adam})  to minimize the average cross-entropy between the estimated output probability distribution $P(a_i)$ and the one-hot-encoded vector $g_i$. The loss function is defined as follows:
\begin{equation}
     \Lagr = \frac{1}{N_s} \sum_{i=1}^{n} D(P(a_i),g_i)
\end{equation}
where $P(a_i)$ is obtained using the softmax function, $g_i$ is the one-hot encoded vector of the $i^{th}$ sample, $N_s$ is the number of samples available for training, and $D(P(a_i),g_i)$ is the cross-entropy distance function defined as:

\begin{equation}
     D(P(a_i),g_i) = - \sum_{j=1}^{n} g_{ij} log (P(a_{ij}))
\end{equation}

\subsection{Federated Learning} 
\label{sec:federate_learning}

\begin{figure*}[!t]
\centering
\vspace{-0.5cm}
    \begin{minipage}[b]{0.58\linewidth}
    \includegraphics[width=\linewidth]{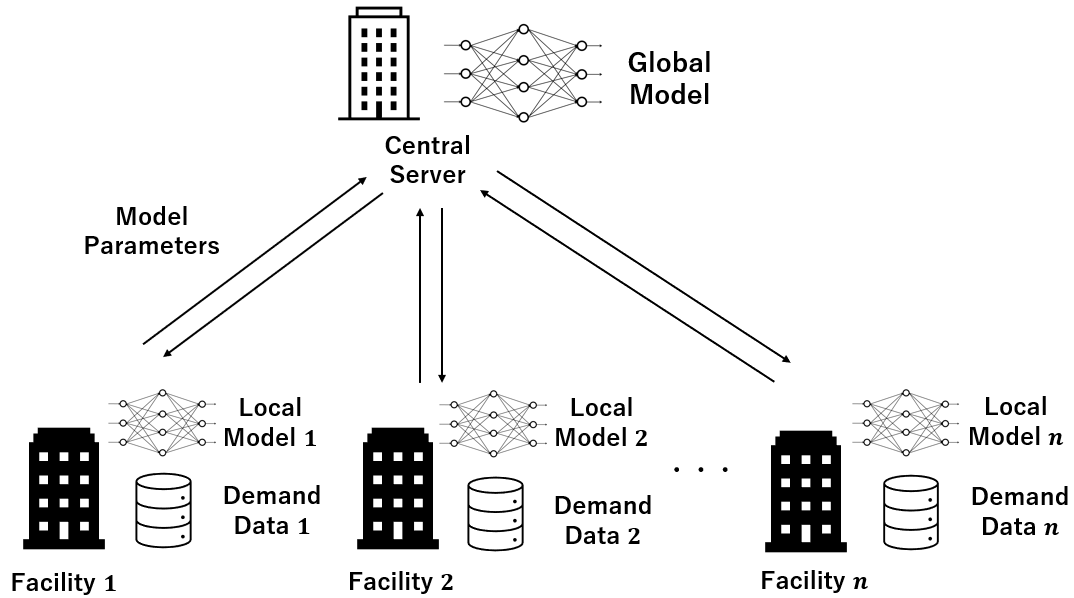}
    \caption{Overview of our federated learning of taxi-demand prediction}
    \label{fig:fl_overview}
    \end{minipage}
    \vspace{-0.5cm}
\end{figure*}

\subsubsection{Our Approach}

Federated learning is a distributed machine learning technique that enables multiple clients to train a model collaboratively without sharing their private data with a central server. In this study, we use the Federated Averaging (FedAvg) algorithm on our federated learning of taxi-demand prediction. FedAvg, proposed by McMahan et al.~\cite{mcmahan2017communication}, is a widely used framework for federated learning due to its simplicity and scalability.

The FedAvg algorithm works as follows: At the beginning of each round, the central server selects a subset of clients to participate in the training process. The server sends the current global model to the selected clients, and each client trains the model using their local data. Specifically, each client updates the model by computing the gradients of their local loss function and performing a gradient descent step. This local update is given by:
\begin{equation}
w_{t+1}^k \leftarrow w_t - \eta g_k
\end{equation}
where $w_t$ and $w_{t+1}^k$ are the model parameters at round $t$ and $t+1$ respectively, $k$ is the client ID, $\eta$ is the learning rate, and $g_k$ is the gradient of the local loss function with respect to the model parameters.

After the local updates are completed, each client sends their updated model to the central server. The server then averages all the received models to obtain a new global model. The global update is given by:
\begin{equation} \label{eq:FedAVG}
w_{t+1} \leftarrow \sum_{k=1}^K\frac{n_k}{n}w_{t+1}^k
\end{equation}
where $n_k$ is the number of data samples held by client $k$, $n$ is the total number of data samples in the system, and $K$ is the total number of clients participating in the training process.

The FedAvg algorithm repeats the above process for a specified number of rounds until convergence. The global model of the server is the final output.

Fig.\ref{fig:fl_overview} shows the overview of our federated learning approach for taxi-demand prediction. Each client represents a specific facility and has access to its own private data. We implemented our approach using PyTorch, a popular machine learning framework, and Flower~\cite{beutel2020flower}, a federated learning framework for PyTorch.

Our approach involves the following steps:
\begin{enumerate}
\item The central server sends the current global model to a subset of clients.
\item Each client trains the model using their local data, and updates the model using the FedAvg algorithm.
\item Each client sends their updated model back to the central server.
\item The central server averages all the received models to obtain a new global model.
\item The above steps are repeated until convergence.
\end{enumerate}

\begin{table}[!tbp]
    \centering
    \caption{Hyperparameters of experiment settings.}
    \begin{tabular}{c|c} \hline
        Criteria & Value (\textbf{bold} default) \\
        \hline
        \textit{Number of prediction classes} & \textbf{4} \\
        \textit{Number of global epochs} & \textbf{300} \\
        \textit{Patience of early stopping} & 10, \textbf{30}, $\infty$ \\
        \textit{Number of facilities} & 4, 8, \textbf{16} \\
        \textit{Number of local epochs} & \textbf{1} \\
        \hline
    \end{tabular}
    \label{tab:setting}
\end{table}

\section{Evaluation} \label{sec:evaluation}

This section describes experimental evaluations. 
Firstly, data collection is described. 
Then, the evaluations of the taxi-demand prediction model described in Section~\ref{sec:TheSystemDetails} and the privacy are described.

\subsection{Data Collection and Setting} 

\subsubsection{Data Collection}
We gathered real-world data from $16$ service facilities in Japan over a period of six months. The collected data includes (1) vehicle information and their trajectories (including idle time), and (2) spatiotemporal data of each customer's pickup and drop-off event for each vehicle. The system determined the trajectory of each customer's trip by merging the two datasets using the vehicle ID and time as the key factors. This resulted in 15,178 trips, with taxi demands ranging from 0 to 20, calculated using a grid size of 1 km and a time slot of 1 hour.

The trajectory data was obtained through GPS for latitude and longitude, with data acquisition intervals of approximately every 5 seconds, with some missing data. To determine the locations of pickup and drop-off events, we used data on vehicle positions during the 45 seconds before and after the event, if available. If the data was not present, the event was omitted from the evaluation data. The number of demands with determined locations and times was 10327. 

\subsubsection{Experimental Setting}
We describe each setting below. 

\paragraph{Data Splitting}
In the following experiments, we split the entire data into three subsets, i.e., 64\% for training data, 16\% for validation data, and 20\% for test data. 
The training data is utilized for training the model, the validation data is for early stopping the training, and the test data is for computing the evaluation metrics described later. 
In the case of federated learning, each facility has the training and validation data, and a central server has the test data. 
We then utilize the split dataset for two models, i.e., a single model and federated learning. 
Each model is trained in the same setting as Section~\ref{sec:eval_taxi-demand}.

\paragraph{Metrics}
We focus on two metrics, i.e., accuracy and balanced accuracy~\cite{brodersen2010balanced} for taxi-demand prediction evaluation. 
Since the gathered data described above are class-imbalanced, the evaluation of the conventional accuracy for prediction results is insufficient. 
Therefore, we adopt the balanced accuracy, which is the average of the accuracy between all the classes. 
We utilize the existing implementations of the scikit-learn library for the above metrics.

\paragraph{Hyperparameters}
Hyperparameters in the experiments are shown in Table~\ref{tab:setting}, where four prediction classes are defined as \textit{non}, \textit{low}, \textit{med}, and \textit{high}. 
We also set the `margin' described in Section~\ref{sec:federate_learning} as 1. 


\subsection{Evaluation of Taxi-Demand Prediction} 
\label{sec:eval_taxi-demand}

\begin{figure}
\vspace{-0.5cm}
\centering
\includegraphics[width=0.9\linewidth]{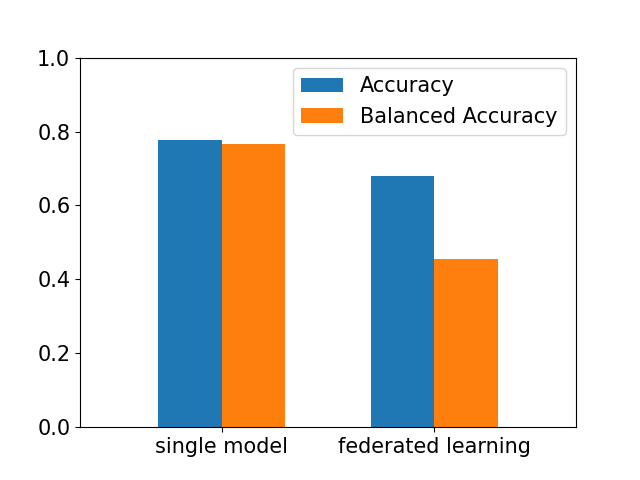}
    \caption{Results of taxi-demand prediction for single model and federated learning.}
    \label{fig:single_federated}
      \hspace{0.1cm}
      \vspace{-0.3cm}
\end{figure}

\begin{figure}[tb]
\centering
\vspace{-0.5cm}
\includegraphics[width=0.9\linewidth]{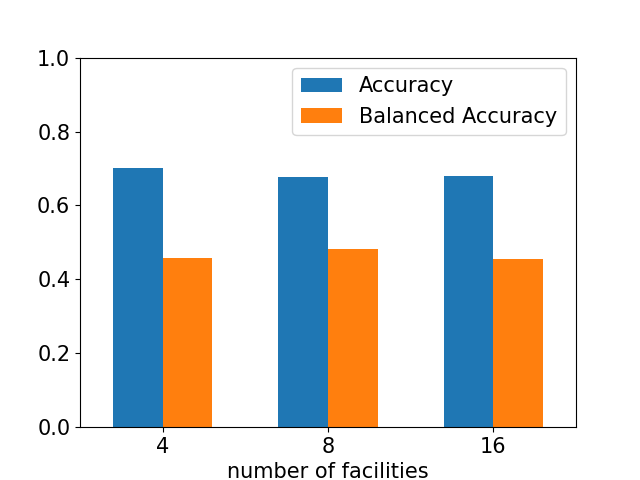}
    \caption{Effect of changing the number of facilities (nodes) in federated learning.}
    \label{fig:facility_num}
      \hspace{0.1cm}
      \vspace{-0.5cm}
\end{figure}

Figure~\ref{fig:single_federated} and Figure~\ref{fig:facility_num} illustrate the comparison between the single model and the proposed federated learning approach, and it shows that the accuracy and balanced accuracy of the federated learning are slightly lower than those of the single model by  0.096, and  0.310, respectively. However, it is essential to highlight that federated learning enables privacy preservation by training the model on decentralized data without compromising the security of the data. This aspect is particularly important for commercial applications, e.g., taxi-demand prediction based on customers' data. Therefore, the slight tradeoff between accuracy and privacy in federated learning is a reasonable compromise, and it makes this approach a practical and promising solution for privacy-sensitive scenarios. 
Specifically, federated learning ensures compliance with privacy regulations such as the General Data Protection Regulation (GDPR) by keeping the data local and not transmitting it to a central server. 


\begin{figure}[tb]
\vspace{-0.5cm}
\centering
\includegraphics[width=0.9\linewidth]{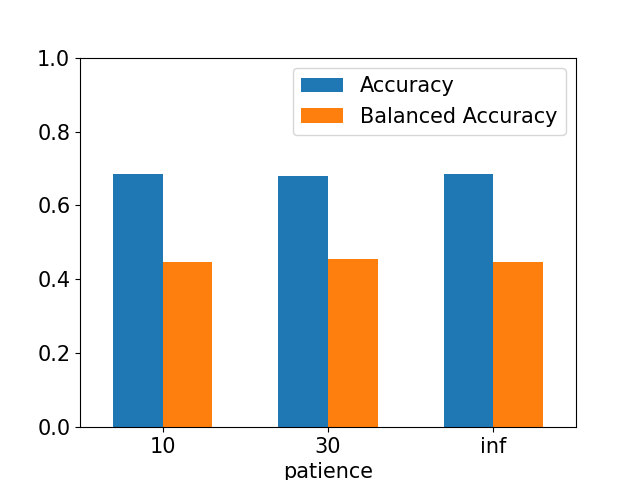}
    \caption{Results for different patience values in federated learning.}
    \label{fig:patience}
      \hspace{0.1cm}
\vspace{-0.5cm}
\end{figure}

Figures~\ref{fig:patience} shows the result of the patience parameter that controls early stopping. This parameter represents the number of epochs required before terminating the training process when no performance improvement is obtained. According to the figure, the system accuracy seems to reach the optimal model with as low as only 10 epochs.

\section{Conclusion} \label{sec:conclution}

In this paper, we presented a novel approach to privacy-preserving taxi demand prediction using federated learning. 
Our proposed system leverages the FedAvg federated learning technique to train a taxi-demand prediction model without compromising the privacy and security of customer data owned by  taxi-providing facilities. 
By enabling facilities to build models on data they would otherwise be unable to access, our approach offers significant benefits in terms of data availability.
To evaluate the effectiveness of our proposed system, we conducted experiments using real-world data collected from 16 taxi service providers in Japan over a period of six months. The results demonstrated that the system accurately predicts demand levels with less than a 1\% decrease in accuracy compared to classical solutions. 

\section*{Acknowledgment}
This work was supported by JST, CREST Grant JPMJCR21M5, Japan, and JSPS, KAKENHI Grant  22K12011, and NVIDIA award.


\end{document}